\documentclass[10pt,twocolumn,letterpaper]{article}

\usepackage{cvpr}              %
\usepackage{multirow}
\usepackage{multicol}
\usepackage{makecell}
\usepackage{subcaption}

\definecolor{cvprblue}{rgb}{0.21,0.49,0.74}
\usepackage[pagebackref,breaklinks,colorlinks,allcolors=cvprblue]{hyperref}
\usepackage[accsupp]{axessibility}
\def\paperID{} %
\def\confName{CVPR}
\def\confYear{2026}

\title{FluidGaussian: Propagating Simulation-Based Uncertainty Toward Functionally-Intelligent 3D Reconstruction
}
\newcommand*\samethanks[1][\value{footnote}]{\footnotemark[#1]}
\author{
Yuqiu Liu$^1$\thanks{The first two authors contributed equally.} , Jialin Song$^1$\samethanks , Marissa Ramirez de Chanlatte$^2$, Rochishnu Chowdhury$^2$,\\Rushil Paresh Desai$^3$, Wuyang Chen$^1$, Daniel Martin$^2$, Michael W. Mahoney$^{2,3,4}$\\
$^1$Simon Fraser University\quad 
$^2$Lawrence Berkeley National Lab\quad 
$^3$University of California, Berkeley\\
$^4$International Computer Science Institute\\
}

\begin{document}

\maketitle

\begin{abstract}
Real objects that inhabit the physical world follow physical laws and thus behave plausibly during interaction with other physical objects.
However, current methods that perform 3D reconstructions of real-world scenes from multi-view 2D images
optimize primarily for visual fidelity, i.e., they train with photometric losses and reason about uncertainty in the image or representation space.
This appearance-centric view overlooks body contacts and couplings,
conflates function-critical regions
(e.g., aerodynamic or hydrodynamic surfaces)
with ornamentation, and 
reconstructs
structures suboptimally, even when physical regularizers are added.
All these can lead to unphysical and implausible interactions.
To address this, we consider the question:
How can 3D reconstruction become aware of real-world interactions and underlying object functionality, beyond visual cues?
To answer this question, we propose FluidGaussian, a plug-and-play method that tightly couples geometry reconstruction with ubiquitous fluid-structure interactions to assess surface quality at high granularity.
We define a simulation-based uncertainty metric induced by fluid simulations and integrate it with active learning to prioritize views that improve both visual and physical fidelity.
In an empirical evaluation on NeRF Synthetic (Blender), Mip-NeRF 360, and DrivAerNet++, our FluidGaussian method yields up to +8.6\% visual PSNR (Peak Signal-to-Noise Ratio) and -62.3\% velocity divergence during fluid simulations.
Our code is available in \href{https://github.com/delta-lab-ai/FluidGaussian}{https://github.com/delta-lab-ai/FluidGaussian}.
\end{abstract}

\section{Introduction} 

3D reconstruction aims to recover the geometry and appearance of real-world 3D scenes from multi-view 2D images. 
Modern radiance-field techniques aim to learn continuous scene representations with strong visual fidelity and efficient rendering.
This includes Neural Radiance Fields (NeRFs) optimized through differentiable volume rendering~\cite{mildenhall2021nerf} and 3D Gaussian Splatting with explicit Gaussian primitives for real-time rendering~\cite{kerbl20233d}. 
Building on these representations, next-best-view (NBV) selection improves reconstruction with limited views by choosing poses that reduce model uncertainty or that maximize Fisher information~\cite{pan2022activenerf,jiang2023fisherrf}. 
Despite such progress, \emph{most pipelines still optimize almost exclusively for visual fidelity}.

\begin{figure}[t]
    \centering
    \includegraphics[width=0.47\textwidth]{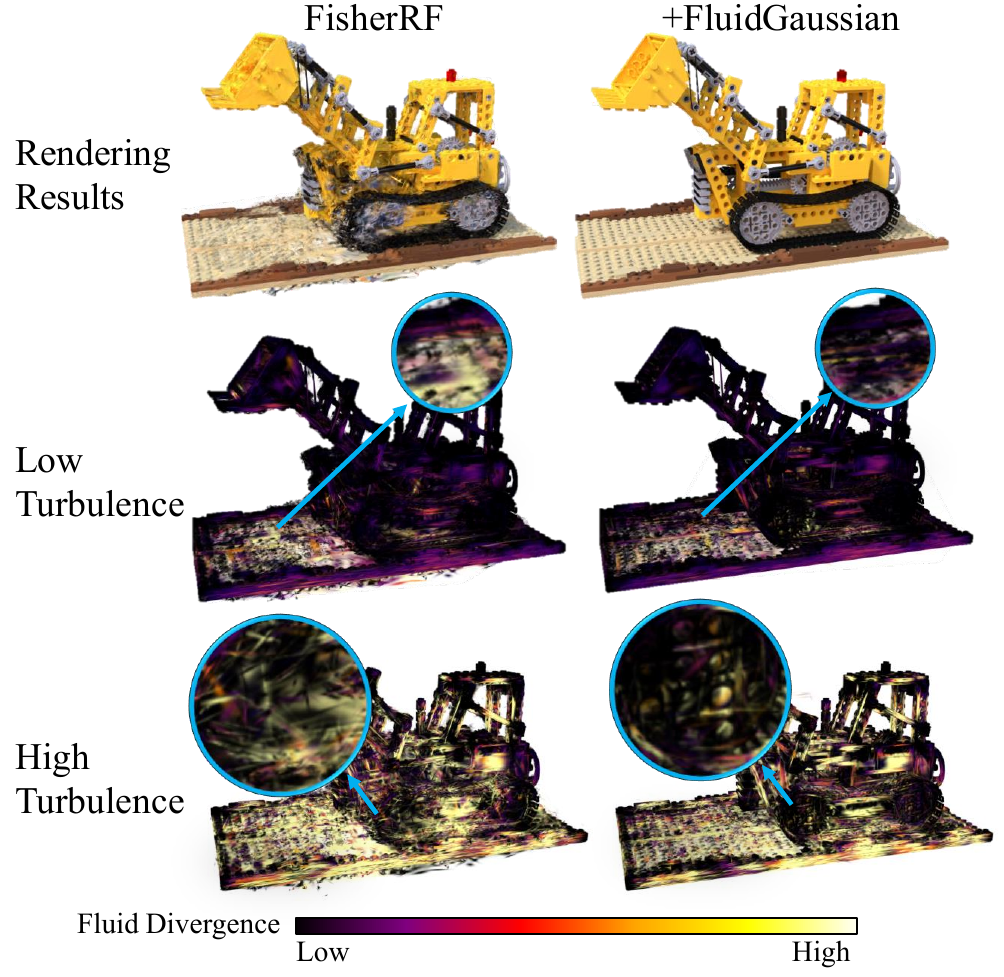}
    \captionsetup{font=small}
\vspace{-1.3 em}
    \caption{When only visual fidelity is optimized and physical awareness is excluded~\cite{pan2022activenerf,jiang2023fisherrf} (left column), reconstructions \emph{may look correct visually} (top), and yet \emph{can fail functionally}: fluid simulation exhibits excessive velocity-field divergence (defined in Section~\ref{sec:method_fluid_uncertainty_metric}), a deficiency that worsens from low to high turbulence (from middle to bottom). With FluidGaussian (right column), both visual fidelity and simulation fidelity are better preserved.
    }
    \label{fig:teaser}
    
\vspace{-1.5 em}
\end{figure}

While visual fidelity is important, in many applications, in particular those in scientific and engineering domains, the goal is not only to reproduce views but also to recover geometry that exists in the real world and that behaves plausibly under interactions~\cite{li2025scorehoi,fieraru2020three}.
Without physics awareness, reconstructed shapes \emph{may visually look correct}, and yet \emph{can fail functionally as 3D assets} (Fig.~\ref{fig:teaser}). 
This inability of existing methods can arise for several reasons: 
\textbf{(i)} they rely only on \emph{appearance cues}, ignoring body contacts and mechanical couplings that govern interactions; 
\textbf{(ii)} they ignore \emph{functional preservation}, treating function-critical regions (such as aerodynamic or hydrodynamic surfaces) the same as cosmetic geometry; and 
\textbf{(iii)} they miss \emph{fine-scale details} that are essential for stable physical behavior. 
Although recent studies examine simulation-related uncertainty~\cite{hodgkinson2023monotonicity,hansen2023learning,mouli2024using,daniels2025uncertainty}, these works focus on scientific machine learning rather than 3D reconstruction. 
Incorporating physics awareness supports ``functional intelligence’’ in 3D vision, potentially enabling assets that interact stably in simulation.
Motivated by these gaps, we ask:

\vspace{-0.5em}
\begin{center}
\fbox{
\parbox{0.9\linewidth}{
\emph{How can 3D reconstruction become aware of real-world interactions and underlying object function, beyond visual cues?}
}
}
\end{center}
\vspace{-0.5em}

In this work, we introduce FluidGaussian, a method to perform physics-aware reconstruction by propagating uncertainty from physical interactions. 
We use fluid simulation as the interaction model, as fluid-structure responses can be used to provide detailed signals of geometry quality (Section~\ref{sec:motivation_fluid_interaction}).
We propose the following methods:
\begin{itemize}[leftmargin=*]
\item \emph{Metrics.} We define a simulation-induced uncertainty metric that measures physical plausibility and interaction fidelity by evaluating reconstructed geometry in standard fluid simulations.
\item \emph{Planners.} We couple this uncertainty with NBV planners,
turning view acquisition physics-aware and guiding poses toward regions where fluid–structure behavior reveals weaknesses beyond photometric cues. 
\item \emph{Evaluation.} We evaluate both visual and simulation-grounded fidelity across synthetic, realistic, and scientific datasets, with a focus on \emph{function-critical} areas.
\end{itemize}
FluidGaussian achieves the performance improvements:
\begin{itemize}[leftmargin=*]
\item
\emph{Improved visual appearance} up to +8.6\% PSNR
on Blender, Mip-NeRF 360, and DrivAerNet++, covering both everyday and scientific scenarios.
\item \emph{Improved simulation fidelity}, leading to up to -62.3\% lower velocity-field divergence (more incompressible), with larger gains at higher Reynolds numbers.
\item \emph{High accuracy on function-critical surfaces}, achieving +7.7\% PSNR on flow-interacting regions.
\end{itemize}

\begin{figure*}[t]
	\centering
	\includegraphics[width=0.95\textwidth]{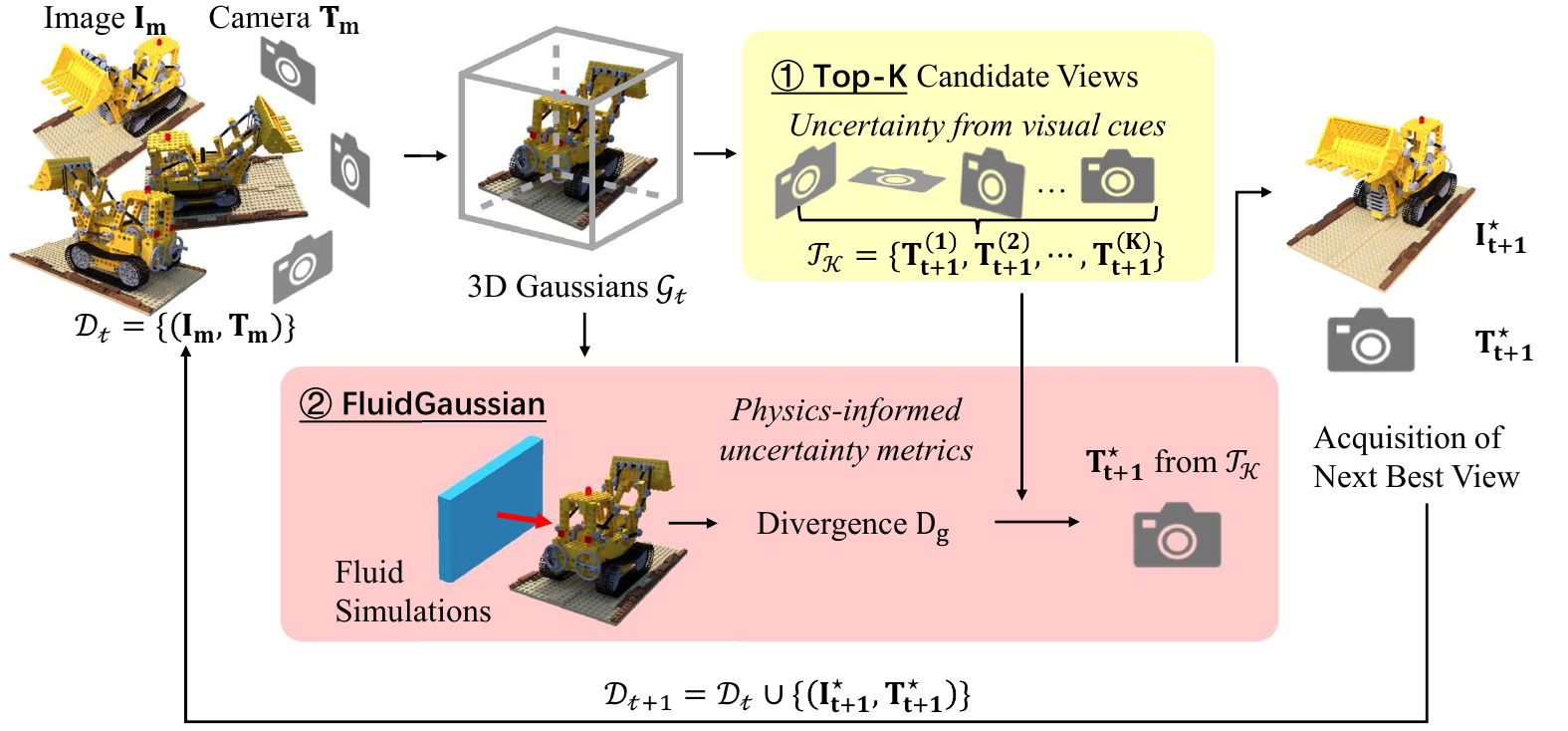}
    \captionsetup{font=small}
\vspace{-0.5 em}
	\caption{Overview of FluidGaussian. To introduce physics-awareness into 3D reconstruction: \textbf{(1)} start from $K$ candidate camera poses proposed by vision-based NBV methods~\cite{pan2022activenerf,jiang2023fisherrf}; and \textbf{(2)} evaluate a physics-informed uncertainty induced from fluid-structure simulations (Section~\ref{sec:method_fluid_uncertainty_metric}) and select the best next camera pose (Section~\ref{sec:method_nbv_selection}). The \textcolor{blue}{blue} block denotes the fluid.}
\label{fig:pipeline}
\vspace{-1.5 em}
\end{figure*}

\section{Background and Motivation}

In this section, we justify the lack of physical quality preservation in standard 3D reconstruction methods, and we detail our plan for quantifying this deficit.

\subsection{Background: Active 3D Reconstruction}
\label{sec:background_active_3d}

\paragraph{Gaussian Splatting.}
For our geometry representation, we consider 3D Gaussian Splatting (3DGS)~\cite{kerbl20233d}.
We parameterize a scene by $N$ anisotropic 3D Gaussians:
\[
\mathcal{G}
=\{g_i\}_{i=1}^N,\qquad 
g_i=(\mathbf{x}_i,\boldsymbol{\Sigma}_i,\alpha_i,\mathbf{c}_i),
\]
where $\mathbf{x}_i\in\mathbb{R}^3$ is the center, $\boldsymbol{\Sigma}_i\in\mathbb{R}^{3\times3}$ the covariance, $\alpha_i\in[0,1]$ the opacity, and $\mathbf{c}_i$ denotes (view-dependent) color coefficients. 
Given calibrated training data $\mathcal{D}=\{(\mathbf{I}_m,\mathbf{T}_m)\}_{m=1}^{|\mathcal{D}|}$ ($\mathbf{I}$ the image view; $\mathbf{T} = [\mathbf{R | \mathbf{t}}]$ the camera pose with the rotation-translation pair), we render view $m$ with a differentiable rasterizer
$
\hat{\mathbf{I}}_m=\mathcal{R}(\mathbf{T}_m;\mathcal{G}),
$
and we learn parameters
$\mathcal{G}$
by minimizing a photometric objective,
$
\min_{\mathcal{G}}\;\mathcal{L}(\mathcal{G})
=\sum_{m=1}^{|\mathcal{D}|}\ell\!\left(\hat{\mathbf{I}}_m,\mathbf{I}_m\right),
$
with a standard per-pixel loss $\ell$ (e.g., SSIM).

\vspace{-1.2 em}
\paragraph{Active View Acquisition.}
Dense image capture for 3D reconstruction can be costly, especially for large or outdoor assets that can rely on UAV (unmanned aerial vehicle) photogrammetry (adding flight-planning, longer flight times, and heavier data processing burdens)~\cite{liu2022learning, ortiz2022isdf, hardouin2020next, chen2024gennbv}.
To reduce effort, active reconstruction lowers capture cost by selecting NBVs using uncertainty or information-gain criteria~\cite{pan2022activenerf,jiang2023fisherrf}, acquiring only the views that most improve reconstruction under a fixed budget.

\subsection{Motivation: Fluid Simulation Exposes Physical Flaws in 3D Reconstructions}
\label{sec:motivation_fluid_interaction}

\paragraph{Visual Overfitting Underestimates Physical Quality.}
Reconstructions \emph{trained only with visual losses} can score highly on image metrics (e.g., PSNR) but \emph{behave poorly in physical simulation}.
This is illustrated in Fig.~\ref{fig:teaser}, where the appearance-optimized baseline renders plausibly, but produces flow fields with large (worse) velocity divergence (defined in Section~\ref{sec:method_fluid_uncertainty_metric}), clear evidence of non-physical behavior, and discrepancy that worsens in high turbulence regimes, where simulation errors amplify.
This is but one example of visual overfitting systematically underestimating true physical quality: reconstructions that score well in pixel-space metrics are revealed as seriously deficient when evaluated via interaction-based simulation.

\vspace{-1.2 em}
\paragraph{Quantification of Physical Quality via Interactions.}
Physical quality can be assessed by how geometry behaves under \emph{interactions}.
For example, with digital twins, models must not just reproduce appearance, but instead they must also predict real behavior under loads and environments~\cite{torzoni2024digital, ferrari2024digital, kapteyn2020toward, willcox2024role, chaudhuri2023predictive}.
This is particularly important when the digital twin makes decisions to update its actions or data collection policies~\cite{ferrari2024digital, willcox2024role}.
In addition, for immersed objects (ones that are partially or fully surrounded by fluids), functional outcomes (drag, lift, stability, aeroacoustics) depend on accurate fluid–structure coupling~\cite{afridi2024fluid}; thus, reconstruction must be simulation-ready and physically consistent, not merely photorealistic surfaces.

\vspace{-1.2 em}
\paragraph{Why Do We Interact with Fluids?}

Beyond being a physical phenomenon of interest, \emph{fluids} can be used to ``probe'' models to expose flaws in 3D reconstruction.
We quantify physical quality by simulating fluid-geometry interactions on the reconstructed surface, rather than relying on sparse rigid-body contacts. 
There are three reasons for this.
\emph{First}, surface coverage and granularity: unlike pointwise rigid-body contacts, fluids interact with the \emph{entire} exposed surface, probing fine geometric defects (thin parts, gaps, occlusions) with higher sensitivity to small shape differences.
\emph{Second}, robust particle-wise metrics: measurements
aggregated over large fluid particle sets provide stable, meaningful estimates, unlike sparse contact events.
\emph{Third,} ubiquity and practical relevance: fluid interactions (air or water) are pervasive in real settings and well supported by standard simulators, making them ubiquitous, and often more informative, than rigid-body interactions.

\vspace{-1.2 em}
\paragraph{Basics of Fluid Simulation.}
Fluid simulations describe how unknown fields $u(\mathbf{x},t)$ (e.g., mass, velocity, pressure) vary over space $\mathbf{x}\in\Omega$ and time $t$.
Governing equations describing fluid evolution typically enforce conservation of mass and momentum,
together with initial and boundary conditions.
Fluid simulations numerically approximate $u$ by discretizing space (finite differences / volumes / elements) and time (explicit or implicit steps), then advancing the solution via algebraic updates.
A common incompressible model evolves velocity $\mathbf{v}$ and pressure $p$ using an advection-diffusion-forcing step followed by a projection that enforces an incompressibility constraint on the velocity field ($\nabla\!\cdot\mathbf{v}=0$):
(i) update a provisional velocity $\mathbf{v}^\ast$ by advection and diffusion with non-pressure forces (e.g., gravity);
(ii) project the provisional velocity onto a space of divergence-free velocity by solving a Poisson equation to correct the flow $\nabla^2 p = \frac{\rho}{\Delta t}\nabla\!\cdot \mathbf{v}^\ast$;
(iii) correct the velocity field by subtracting the gradient of the pressure field $\mathbf{v}^{n+1}=\mathbf{v}^\ast-\frac{\Delta t}{\rho}\,\nabla p$; and
(iv) apply appropriate boundary conditions.
These steps constitute a consistent and stable algorithm which yields physically plausible flows, and they can be extended to compressible or multiphase settings by modifying the governing terms and constraints.
Beyond this pipeline, recent works also studied learning-based simulators~\cite{hansen2023learning,chen2024data,olivares2024clover,benitez2025neural,xu2025hybrid,utkarsh2025end,ma2026learning}.

\section{The FluidGaussian Method}

\subsection{Overview}

\paragraph{Pipeline.}

Our goal is to preserve both visual and physical quality during 3D reconstruction by incorporating physical interactions beyond visual cues.
Motivated by Section~\ref{sec:motivation_fluid_interaction}, 
we develop FluidGaussian, a physics-driven uncertainty that complements existing vision-based NBV strategies that rely on visual cues.
We summarize the FluidGaussian pipeline in Fig.~\ref{fig:pipeline}.

\vspace{-1.2 em}
\paragraph{Step 1: Vision-based Uncertainty Proposes Top-K Candidate Camera Poses.}
In the context of active 3D reconstruction (Section~\ref{sec:background_active_3d}),
given a finite set of available camera poses $\mathcal{T}$ and a small seed of initial views, our pipeline first proposes a pool $\mathcal{T}_K$ of $K$ candidate poses ($K \ll |\mathcal{T}|$).

We adopt ActiveNeRF~\cite{pan2022activenerf} or FisherRF~\cite{jiang2023fisherrf} as our baseline NBV planners.
We use
$\mathcal{D}_t=\{(\mathbf{I}_m,\mathbf{T}_m)\}_{m=1}^{|\mathcal{D}_t|}$
to denote the views and poses collected up to time step $t$; and denote
the reconstructed geometry (3D Gaussians, Section~\ref{sec:background_active_3d}) as $\mathcal{G}_t=\arg\min_\mathcal{G} \mathcal{L}(\mathcal{G}, \mathcal{D}_t)$. 
ActiveNeRF and FisherRF choose the next pose by maximizing a utility $U$ conditioned on the current state $\mathcal{G}_t$:
$
\mathbf{T}_{t+1}^\star = \arg\max_{\mathbf{T} \in \mathcal{T}\setminus\mathcal{D}_t} U(\mathbf{T}, \mathcal{G}_t)
$.
The utility $U$ is typically an uncertainty quantification metric, e.g., variance reduction for the radiance field distribution~\cite{pan2022activenerf} or Fisher Information in volumetric rendering~\cite{jiang2023fisherrf}, derived from visual cues.

To plug FluidGaussian into ActiveNeRF and FisherRF,
we retain the top-$K$ candidate camera poses proposed by these methods, $\mathcal{T}_K = \{\mathbf{T}_{t+1}^{(1)}, \mathbf{T}_{t+1}^{(2)}, \cdots, \mathbf{T}_{t+1}^{(K)}\}$, rather than selecting only a single candidate,
and we evaluate our physics-informed uncertainty metric rendered at each of K candidate poses, as explained in the next step.

\vspace{-1.4 em}
\paragraph{Step 2: FluidGaussian Selects the Best Physics-Informed Camera Pose.}

Given the reconstructed geometry $\mathcal{G}_t$,
we exploit incompressible fluid simulation to define a geometry-sensitive, simulation-induced uncertainty metric (see Section~\ref{sec:method_fluid_uncertainty_metric}).
Using Smoothed Particle Hydrodynamics (SPH), we ``flush'' the reconstructed object with fluid particles, and we evaluate the simulation quality of near-surface fluid particles.
Unstable or degraded fluid simulation around the near-surface region can expose geometric artifacts (e.g., gaps, self-intersections, or thin/occluded errors).
Our uncertainty metric will select the best camera pose from the top-$K$ candidates proposed in step 1, i.e., $\mathbf{T}_{t+1}^\star \in \mathcal{T}_K$, thus balancing the visual quality and physical quality (see Section~\ref{sec:method_nbv_selection}).

The corresponding NBV is acquired as $\mathbf{I}_{t+1}^{\star}$,
and we then update $\mathcal{D}_{t+1} = \mathcal{D}_t\cup\{(\mathbf{I}_{t+1}^\star,\mathbf{T}_{t+1}^\star)\}$ and
$
\mathcal{G}_{t+1}=\arg\min_{\mathcal{G}}\;\mathcal{L}\!\left(\mathcal{G},\mathcal{D}_{t+1}\right)
$,
and we iterate until the budget is exhausted.

\subsection{Measuring Physical Uncertainty of 3D Reconstruction via Fluid Simulation}
\label{sec:method_fluid_uncertainty_metric}

\begin{figure}[t]
    \centering
    
    \includegraphics[width=0.38\textwidth]{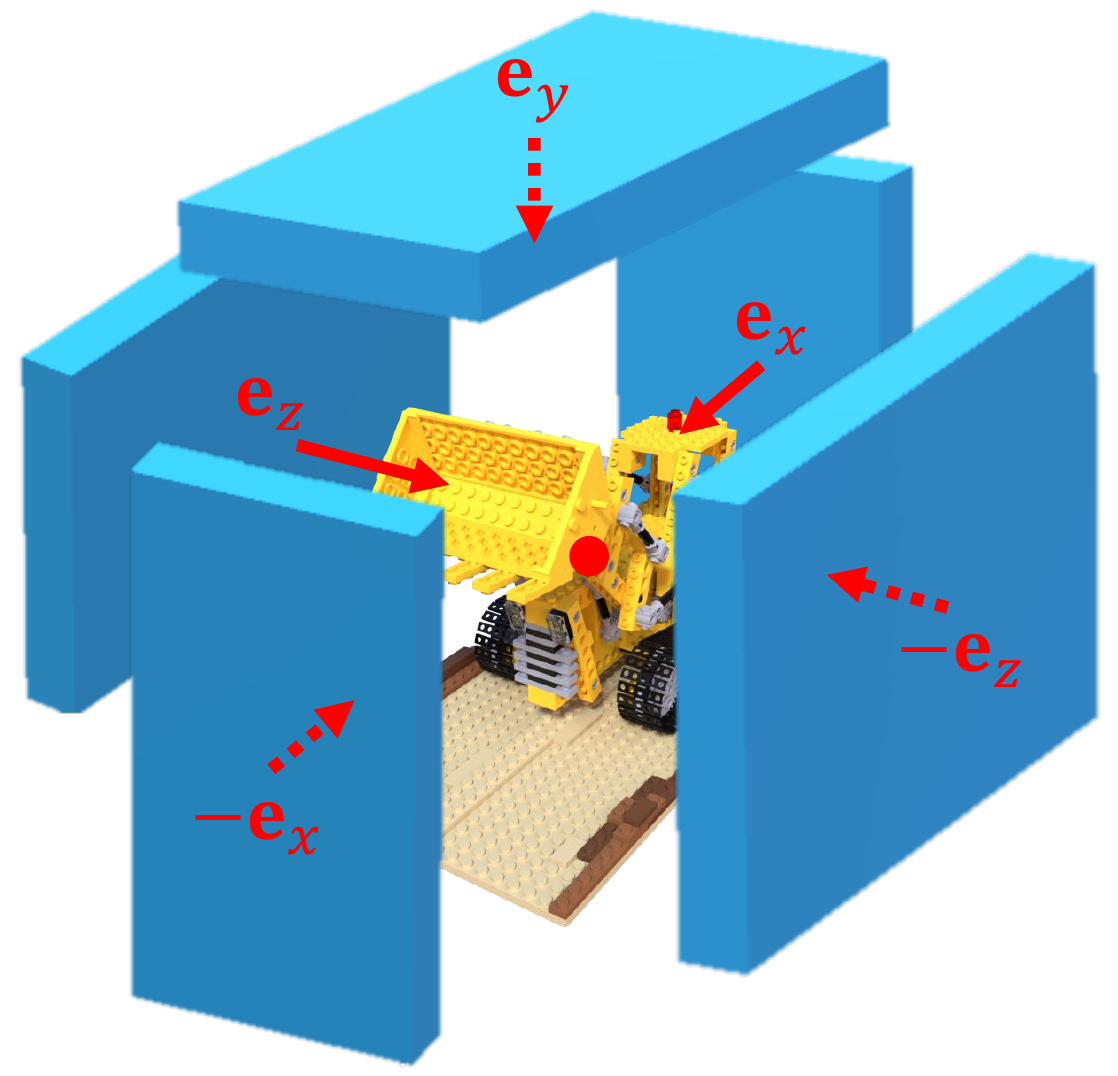}
    \vspace{-1.2 em}
    \captionsetup{font=small}
    \caption{To make our physical uncertainty statistically meaningful, FluidGaussian measures over \emph{individual} fluid simulations from five initial conditions (i.e., five individual simulations). \textcolor{blue}{Blue} blocks denote fluids. We use a Y-up coordinate system.}
    \vspace{-2 em}
    \label{fig:simulation}
\end{figure}
Here, we discuss our simulation-based metric that can quantify the ``physical uncertainty'' of a 3D geometry.\footnote{Prior active-reconstruction methods score views by \emph{visual uncertainty}~\cite{pan2022activenerf,jiang2023fisherrf}. We should emphasize that our divergence-based metric is \emph{not} a probabilistic uncertainty; it is a simulation-propagated indicator of the reconstructed surface’s \emph{physical reliability}. For simplicity, however, we abuse terminology and refer to it as a physics-informed ``uncertainty.''}

\vspace{-1.2 em}
\paragraph{Setup.} 
To reduce directional bias and improve robustness, we ``release'' fluid from multiple initial conditions and run short rollouts per direction, and we then convert the surface-near divergence over individual simulations into a single score to re-rank NBV candidates.
We use divergence as a view-dependent uncertainty instead of a differentiable loss, because backpropagating divergence through fluid simulators would be prohibitively expensive.
 
Specifically, we run individual SPH simulations from five initial conditions ($IC$) per uncertainty measurement: fluid is splashed toward the reconstructed object from four horizontal directions ($\mathbf{e}_x, -\mathbf{e}_x, \mathbf{e}_z, -\mathbf{e}_z$) and from the top ($-\mathbf{e}_y$) with a fixed initial speed (here we use Y-up coordinate system).
Here we omit the bottom-up direction since cameras are rarely placed below the object and bottom-facing surfaces seldom affect functionally important regions.
This is illustrated in Fig.~\ref{fig:simulation}. Each rollout uses a small time step $\Delta t_\text{sim}$ for $T_\text{sim}$ steps and treats the current 3DGS as a solid boundary via an on-the-fly voxel conversion to impose no-penetration constraints.

\vspace{-1.2 em}
\paragraph{DFSPH.}
We adopt the \emph{Divergence-Free Smoothed Particle Hydrodynamics} (DFSPH) scheme~\cite{bender2015divergence}, which predicts intermediate velocities by advection and diffusion, ignoring pressure, and then applies an iterative pressure projection to enforce two key physical constraints: (i) density consistency $\rho\!\approx\!\rho_0$; and (ii) zero velocity divergence $\nabla\!\cdot\!\mathbf{v}\!\approx\!0$. 
The latter condition, often called the \emph{divergence-free} condition, ensures the simulated flow is incompressible, i.e., local fluid volumes are preserved and no artificial mass compression or expansion occurs. This is the appropriate regime for low Mach number flows, in which fluid velocities are small relative to the speed of sound.
DFSPH is efficient,
making it suitable as a plug-and-play physics probe during 3D reconstruction.

\vspace{-1.2 em}
\paragraph{Divergence Estimator.}

Although we are using a divergence-free SPH, the simulation becomes inaccurate at the fluid-structure interface, leading to numerical errors, indicating that a worse fluid-structure interface may lead to a higher divergence~\cite{bender2016divergence,bao2017immersed,beltman2017conservative,yildiran2024pressure}.

We estimate velocity divergence at a fluid particle $i$ by an SPH-weighted sum over neighbors within a finite support. Let $(\mathbf{x}_i,\mathbf{v}_i)$ be position and velocity, $V_j$ the volume of neighbor $j$, and $W(\cdot,h_k)$ a smoothing kernel with gradient $\nabla W$. This weighted sum can be interpreted as a local interpolation of the velocity field using the cubic B-spline kernel~\cite{monaghan1985refined}.
With the neighbor set
\(
\mathcal{N}_{h_r}(i)=\{\,j\mid \|\mathbf{x}_i-\mathbf{x}_j\|\le h_r, j\ne i\,\}
\)
(implemented with a uniform grid and a cutoff radius $r$), the discrete divergence is
\begin{equation}
(\nabla\!\cdot\!\mathbf{v})_i
~\approx~
\sum_{j\in\mathcal{N}_{h_r}(i)}
V_j \cdot \,(\mathbf{v}_j-\mathbf{v}_i)\cdot
\nabla W(\mathbf{x}_i-\mathbf{x}_j,\,h_k).
\label{eq:sph-div}
\end{equation}
We use the magnitude as the per-particle measure, $D_i \;=\; \big|(\nabla\!\cdot\!\mathbf{v})_i\big|$, and we consider only fluid particles; non-fluid particles are assigned values of zero.

\begin{figure}[t]
\vspace{-1em}
    \centering
    \includegraphics[width=0.42\textwidth]{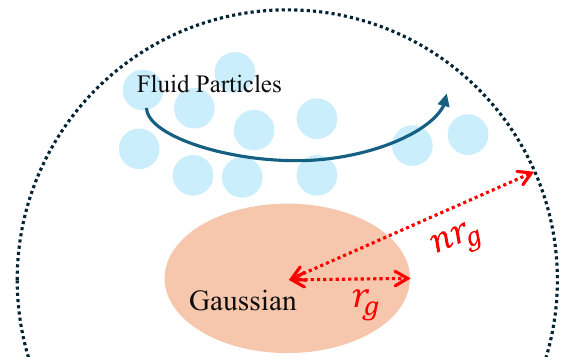}
    \captionsetup{font=small}
    \vspace{-1 em}
    \caption{3D Gaussian accumulates divergence introduced by fluid particles in its neighborhood. Voxelized rigid particles (from 3D Gaussians) are omitted.}
    \label{fig:aggregation}
    
    \vspace{-1em}
\end{figure}

\subsection{Physics-Aware NBV Selection}
\label{sec:method_nbv_selection}

Here, we discuss how to measure our simulation-based uncertainty for each 3D Gaussian, and we then render the per-Gaussian measure into a view from a candidate camera pose $\mathbf{T} \in \mathcal{T}_K$.

\vspace{-1.2 em}
\paragraph{Integration of Divergence into a Per-Gaussian Metric.}
\label{sec:per_g_metric}
For each initial condition $IC\in\{1,\ldots,5\}$, we compute per-particle divergence magnitudes $D_i^{(IC)}=\big|(\nabla\!\cdot\!\mathbf{v})_i^{(IC)}\big|$ using the explicit SPH estimator over neighbors within a finite support.
To associate these simulation quantities to the Gaussian primitives in the 3D representation, we assign to each Gaussian $g$ (centered at $\boldsymbol\mu_g$ with radius $r_g$ as the maximum principal radius of its covariance ellipsoid)
the statistics of fluid particles that pass through its spatial extent during the simulation.
We denote by $\mathcal{P}_g^{(IC)}$ the set of fluid particles located within the sphere (radius $n\cdot r_g$) for initial condition~$IC$:  
{\small
\begin{equation}
    \quad \mathcal{P}_g^{(IC)} ~=~ \big\{\, i\text{-th particle}\ \big|\ \|\mathbf{x}_i^{(IC)}-\boldsymbol\mu_g\|\le n \cdot r_g, i \in [1, N]\big\},
    \label{eq:fluid_particles}
\end{equation}
}%
where $N$ is the total number of fluid particles, and we use $n=3$.
Then we aggregate $D_i^{(IC)}$ to get the divergence around the Gaussian $D_g^{(IC)}$:
\begin{equation}
    \quad D_g^{(IC)} ~=~ \frac{1}{\lvert \mathcal{P}_g^{(IC)}\rvert}\sum_{i\in \mathcal{P}_g^{(IC)}} D_i^{(IC)},
\end{equation}
which is shown in Fig.~\ref{fig:aggregation}.
We then choose one initial condition out of five that introduces the largest divergence,
$
D_g ~=~ \max_{IC \in \{1,2,3,4,5\}} D_g^{(IC)} , %
$
to obtain a single per-Gaussian physics score $D_g$.
Note that this procedure is designed to be metric-agnostic.
For example, one could replace $D_i^{(IC)}$ with vorticity magnitude, yielding a per-Gaussian vorticity score in the same form. %

\vspace{-1.3 em}
\paragraph{View Scoring.} \label{sec:view_scoring}
For a candidate camera $\mathbf{T}$, we convert the per-Gaussian scores $\{D_g\}$ into a single scalar by \emph{visibility-weighted} accumulation.
Let $\mathcal{G}_{\mathbf{T}}$ be the subset of all Gaussians that appear in the rendered image of the view $\mathbf{T}$.
If the renderer provides a screen-space radius $R_g(\mathbf{T})$ for each visible Gaussian, we define a soft area weight:
\begin{equation}
    w_g(\mathbf{T}) ~=~ \frac{\pi\, R_g(\mathbf{T})^2}{H\,W}\ \in [0,1],
\end{equation}
where $H\times W$ is the image resolution (weights are clamped to $[0,1]$ in practice).
The \emph{view score} is then:
\begin{equation}
    S(\mathbf{T}) ~=~ \sum_{g\in \mathcal{G}_{\mathbf{T}}}  w_g(\mathbf{T})\, D_g ,
    \label{eq:view_score}
\end{equation}
with the unweighted sum $S(\mathbf{T})=\sum_{g\in \mathcal{G}_{\mathbf{T}}} D_g$ used when radii are unavailable.
We rank top-$K$ candidate camera poses by $S(\mathbf{T})$ in descending order, and we select the top-1 (i.e., we choose the pose with the largest divergence, suggesting the worst simulation quality and physical uncertainty).

\subsection{Quantifying Function-Critical Region}
\label{sec:method_function_critic}
To analyze how object or scene properties affect reconstruction and simulation outcomes, we introduce quantitative measures of \emph{function-critical regions} for each geometry.
This leads to a subset of test views corresponding to physically meaningful structures, such as the aerodynamic frontal surfaces of the vehicles.

\begin{figure}[b]
    \vspace{-0.5 em}
    \centering
    \includegraphics[width=0.47\textwidth]{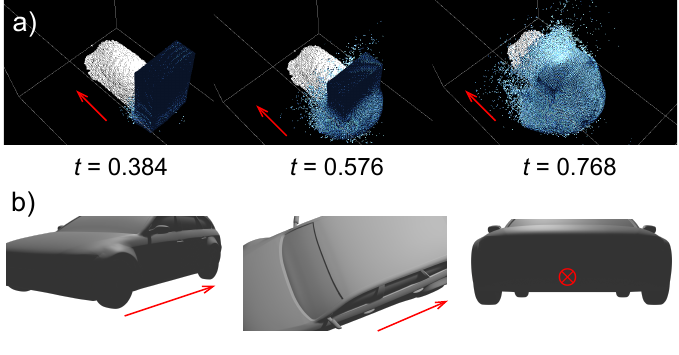}
    \captionsetup{font=small}
    \vspace{-1.6 em}
    \caption{Quantification of function-critical region of DrivAerNet++~\cite{elrefaie2024drivaernet++}: a) Simulation on the $-\mathbf{e}_{x}$ direction to quantify functional-critical regions; b) Examples on function-critical regions in test set. Red arrow shows the direction of flow in the simulation. The rightmost figure indicates the rear of the vehicle.}
    \vspace{-1.2em}
    \label{fig:function_critical_region}
\end{figure}

\begin{figure*}[t]
	\centering
	\includegraphics[width=1.0\textwidth]{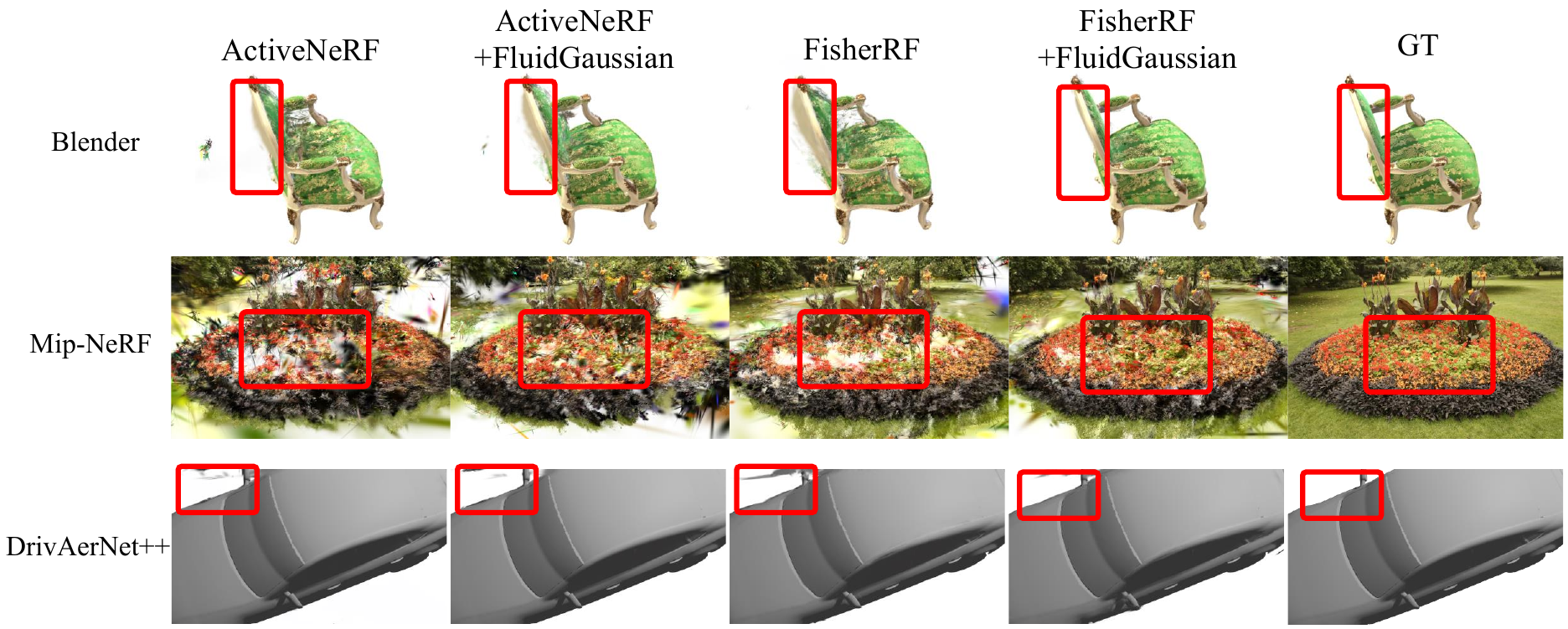}
    \vspace{-2em}
    \captionsetup{font=small}
	\caption{Visualization of the object from Blender~\cite{mildenhall2021nerf}, MipNeRF360~\cite{barron2022mip}, and DrivAerNet++~\cite{elrefaie2024drivaernet++}. Each column represents different methods, from left to right: ActiveNeRF, ActiveNeRF + FluidGaussian, FisherRF, FisherRF + FluidGaussian, ground truth (GT).
    }
    \label{fig:visualization}
\vspace{-0.8em}
\end{figure*}

\begin{table*}[ht]
\centering
\captionsetup{font=small}
\caption{Visual reconstruction results on Blender~\cite{mildenhall2021nerf}, MipNeRF360~\cite{barron2022mip}, and DrivAerNet++~\cite{elrefaie2024drivaernet++} datasets. Adding FluidGaussian improves visual quality in most cases, with the largest gains on DrivAerNet++~\cite{elrefaie2024drivaernet++} when using ActiveNeRF~\cite{pan2022activenerf} as the baseline.}

\vspace{-0.5em}

\resizebox{0.87\linewidth}{!}{
\begin{tabular}{lccccccccc}
\toprule
\multirow{2}{*}{Method} &
\multicolumn{3}{c}{(a) Blender} &
\multicolumn{3}{c}{(b) MipNeRF360} &
\multicolumn{3}{c}{(c) DrivAerNet++} \\ %
\cmidrule(lr){2-4}\cmidrule(lr){5-7}\cmidrule(lr){8-10} %
& PSNR$\uparrow$ & SSIM$\uparrow$ & LPIPS$\downarrow$ &
PSNR$\uparrow$ & SSIM$\uparrow$ & LPIPS$\downarrow$ &
PSNR$\uparrow$ & SSIM$\uparrow$ & LPIPS$\downarrow$  \\ %
\midrule
ActiveNeRF & 21.1525  & 0.8527  & 0.1397  & \textbf{12.9245} & \textbf{0.4331} & \textbf{0.5036} & 17.3840 & 0.9079 & 0.1595  \\
+FluidGaussian & \textbf{21.7642}  & \textbf{0.8613}  & \textbf{0.1230}  & 12.4790 & 0.4248 & 0.5067 & \textbf{18.8727} & \textbf{0.9297} & \textbf{0.1269}  \\ %
\midrule 
FisherRF   & 23.4230  & 0.8761  & 0.1070  & 15.3237 & 0.4704 & 0.4561 & 20.1127 & 0.9339 & 0.1250  \\ %
+FluidGaussian & \textbf{24.7433}  & \textbf{0.8907}  & \textbf{0.0923}  & \textbf{15.5502} & \textbf{0.4717} & \textbf{0.4522} & \textbf{21.0733} &  \textbf{0.9462} & \textbf{0.1047} \\ %
\bottomrule
\end{tabular}
}
\label{tab:act_fisher_synth_real_blend_drive}
\vspace{-1.5em}
\end{table*}
To approximate the dominant aerodynamic interactions in realistic driving scenarios, simulated particles are emitted towards the vehicle’s front surface, as illustrated in Fig.~\ref{fig:function_critical_region}(a). 
For each Gaussian primitive $g$, we accumulate a particle interaction count indicating the number of particles that collide with or pass through its spatial influence region.
Following the notation in Eq.~\ref{eq:fluid_particles}, the count of interacting particles is denoted as
$|\mathcal{P}_g|$.
This count indicates the strength of local fluid interaction, where larger values correspond to more physically active or exposed regions. 

For view evaluation, we adopt the \textit{view score} formulation from Eq.~\ref{eq:view_score}, computing the value $C(\mathbf{T})$ for each test view $\mathbf{T}$:
\begin{equation}
C(\mathbf{T}) ~=~ \sum_{g\in \mathcal{G}_{\mathbf{T}}}  w_g(\mathbf{T})\, |\mathcal{P}_g| .
\label{eq:count_view_score}
\end{equation}
This formulation remains consistent with previous metrics while grounding the view evaluation in physically simulated evidence, thus favoring views that contain surfaces with stronger frontal-flow interaction, such as leading edges and frontal structures. Examples are shown in Fig.~\ref{fig:function_critical_region}(b).

\section{Empirical Evaluation}

In this section, we describe the results of our empirical evaluation of FluidGaussian.

\subsection{Implementation Details}

\paragraph{Datasets.} 
We evaluate on three benchmarks: (a) Blender (synthetic scenes with photorealistic renders)~\cite{mildenhall2021nerf}; (b) MipNeRF360 (real, forward-facing scenes with a large field of view)~\cite{barron2022mip}; and (c) DrivAerNet++ (a parametric 3D car dataset designed for aerodynamic studies)~\cite{elrefaie2024drivaernet++}.
For (a) and (b), we follow the original train-test splits and evaluation scripts.
Due to the unbounded spatial scale of MipNeRF360~\cite{barron2022mip} scenes, we normalize the domain before simulation.
For (c), preprocessing details are provided in the Appendix~\ref{supp:detail}. We report PSNR$\uparrow$, SSIM$\uparrow$, and LPIPS$\downarrow$.

\vspace{-1.4 em}
\paragraph{Setup.} 
Active view acquisition begins from 2 biased views and proceeds until a total budget of 10 views is reached. Unless otherwise stated, all methods use the same training schedule, optimizer, and hyperparameters (Table \ref{tab:train-sim}).
During simulation, we set the initial fluid velocity to 3 for the four horizontal directions, while using 0 for the top-down direction, where gravity drives the fluid motion.
We voxelize Gaussians into particles using NVIDIA Kaolin~\cite{jatavallabhula2019kaolin} and perform both simulation and divergence computation in particle space, which allows the resulting divergence to be directly aggregated to Gaussian centers and rendered into view scores.
Note that for fluid simulation, we adopt the properties of water at room temperature. 
Additional details are available in the Appendix~\ref{supp:detail}.

\vspace{-1.2 em}
\paragraph{Baseline Methods.} 
We compare against two widely used uncertainty-driven NBV baselines:
ActiveNeRF~\cite{pan2022activenerf}; and FisherRF~\cite{jiang2023fisherrf}. 
Their original selection heuristics are kept intact; our method plugs into the same training loop and budget for a fair comparison. 
Following~\cite{jiang2023fisherrf}, to make a fair comparison, ActiveNeRF is re-implemented with 3D Gaussian Splatting.

\begin{table}[t]
\centering
\small
\setlength{\tabcolsep}{2pt}
\captionsetup{font=small}
\caption{Training and simulation settings for all datasets.}
\vspace{-0.5 em}
\resizebox{0.47\textwidth}{!}{
\begin{tabular}{l c c c }
\toprule
 Dataset &  Blender & MipNeRF360  & DrivAerNet++ \\
\midrule
\multicolumn{4}{l}{\textbf{Training}} \\
\midrule
Resolution & $800\times800$ & $1600\times1040$ & $1600\times900$ \\
Test views & 200 & 87 & 100 \\
Opacity ($\alpha$) threshold & 0.3 & 0.4 & 0.25 \\
Iterations  & \multicolumn{3}{c}{30000} \\
Initial views & \multicolumn{3}{c}{2} \\
View budget & \multicolumn{3}{c}{10} \\
Time for training & $\sim$2h & $\sim$3-4h & $\sim$2h \\
GPU memory  & \multicolumn{3}{c}{$\sim$5 GB} \\
\midrule
\multicolumn{4}{l}{\textbf{Simulation}} \\
\midrule
Fluid/rigid density & \multicolumn{3}{c}{1000 / 2200 (kg/m$^3$)} \\
Simulation steps & \multicolumn{3}{c}{750} \\
$\Delta t_{\text{sim}}$ & \multicolumn{3}{c}{0.002 s} \\
Fluid dynamic viscosity & \multicolumn{3}{c}{10 (kg/(m$\cdot$s))} \\
Velocity & \multicolumn{3}{c}{3 m/s} \\
Particle radius & \multicolumn{3}{c}{0.025 m} \\
Cutoff radius $h_r$ & \multicolumn{3}{c}{0.1} \\
Kernel size $h_k$ & \multicolumn{3}{c}{0.1} \\
Num. fluid particles $N$  & 50k-150k & 15k-30k  & 50k-150k \\
Domain size  & [-1.5, 1.5] & [0, 1]  & [-1.5, 1.5] \\
Simulation time  & $\sim$2.5 min & $\sim$3-6 min & $\sim$2.5 min \\
GPU memory & $\sim$4 GB & $\sim$3-10 GB & $\sim$4 GB\\
GPU type & \multicolumn{3}{c}{RTX 6000 Ada; A100-PCIE-40GB} \\
\bottomrule
\end{tabular}
}
\label{tab:train-sim}

\vspace{-1.5 em}
\end{table}

\subsection{FluidGaussian Improves Visual Quality}

FluidGaussian uses per-Gaussian simulation cues (divergence) to steer NBV toward \emph{physically suspect} regions (e.g. thin parts, contacts, occlusions) that appearance-only policies under-sample, thereby yielding assets with more plausible fluid dynamics.
Under the same 10-view budget (2 biased initial views shared by all methods), augmenting ActiveNeRF and FisherRF with our selector (\textit{+FluidGaussian}) improves reconstruction across Blender, MipNeRF360, and DrivAerNet++: PSNR and SSIM increase while LPIPS decreases (Table~\ref{tab:act_fisher_synth_real_blend_drive}), confirming that physics-aware view selection refines appearance (Fig.~\ref{fig:visualization}), although visual quality is not the primary target of our FluidGaussian. More quantitative results can be seen in Appendix~\ref{supp:more_quantitative}.

\subsection{FluidGaussian Refines Simulation-Induced Physics Quality}

\label{exp:phy_quality}

Here, we evaluate how well reconstructed assets preserve fluid–structure plausibility.
To do this, we measure the mean fluid divergence when a steady flow passes the object from
five canonical directions $\mathcal{S}_d=\{\mathbf{e}_x, -\mathbf{e}_x, \mathbf{e}_z, -\mathbf{e}_z, -\mathbf{e}_y\}$.
For a given direction $d\!\in\!\mathcal{S}_d$, we run the same particle-based solver used in our training loop, and we report the per-Gaussian mean divergence.
Our final score averages across directions.

\begin{table}[t]
\centering
\captionsetup{font=small}
\caption{Comparison of divergence ($\bar{D}\downarrow$) after optimization. ``Empty Scene'': simulation without a rigid body. ``Perfect Geometry'': geometry trained with at least 20 views. Value differences in empty-scene divergence are caused by different spatial domains.}

\vspace{-0.3 em}
\resizebox{0.43\textwidth}{!}{
\setlength{\tabcolsep}{2pt}
\begin{tabular}{lcccc}
\toprule
Datasets & Blender & MipNeRF360  & DrivAerNet++ \\
\midrule
Empty Scene  &  $5.67\times10^{-5}$  & 0.0010 & $4.34\times10^{-4}$ \\
Perfect Geometry & 0.0058  & 0.0014 & 0.0371 \\
\midrule
ActiveNeRF     & 0.0262 & 0.0130  & 0.0894  \\
+FluidGaussian & \textbf{0.0145}  & \textbf{0.0049}  & \textbf{0.0801}  \\
\midrule
FisherRF       & 0.0144  & 0.0025  & 0.0609  \\
+FluidGaussian & \textbf{0.0124}  & \textbf{0.0022}  & \textbf{0.0608} \\
\bottomrule
\end{tabular}
}

\vspace{-1. em}
\label{tab:div_cfl_comp}
\end{table}
We compare each baseline before and after adding our FluidGaussian module
(“Baseline” vs “+FluidGaussian”) across three benchmarks. We define geometry divergence as $\bar{D} = \frac{1}{|\mathcal{G}|}\sum_{g\in \mathcal{G}}\pi s_g^2 D_g$, where $s_g = \operatorname{tr}\!\left(\boldsymbol{\Sigma}_i\right)$.

As summarized in Table~\ref{tab:div_cfl_comp}, adding FluidGaussian consistently reduces geometry divergence $\bar{D}$
across all datasets and both baselines. 
This result shows that methods based on FluidGaussian produce fewer spurious sources or sinks, indicating that the reconstructed geometry supports better mass-conservation during simulation. We report divergence for two reference settings: an empty scene, which contains no rigid body, and   ``perfect'' geometry, where the object is trained using far more views than in our pipeline. These results serve as baselines for interpreting divergence. The perfect-geometry case provides the lowest divergence that the reconstruction can reach. For the empty scene, different values are reported for each dataset because the spatial domains of the reconstructed objects differ, which changes how the flow field is distributed.

\begin{figure}[t]
    \centering
    \includegraphics[width=0.47\textwidth]{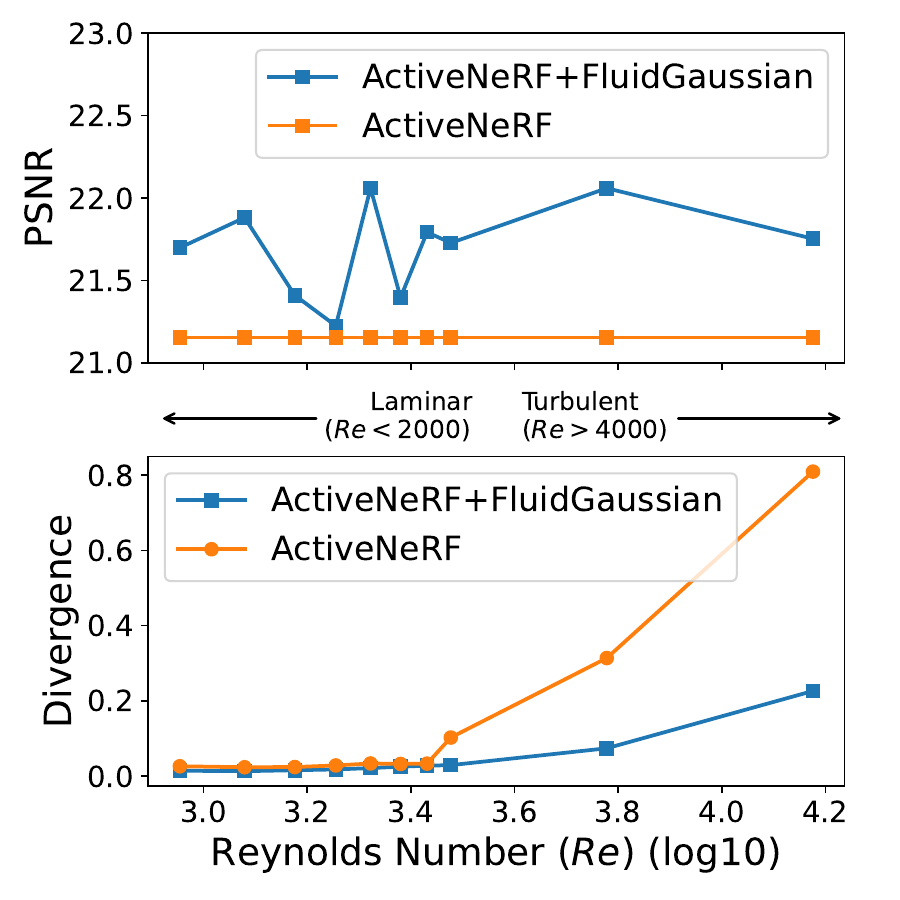}
    
    \vspace{-1.5 em}
    \captionsetup{font=small}
    \caption{PSNR and divergence results on Blender~\cite{mildenhall2021nerf}. 
    FluidGaussian mitigates physical defects increasingly as the fluid transitions from laminar to turbulent. As the fluid velocity rises and the Reynolds number $Re$ increases, appearance-only reconstructions exhibit markedly amplified physical errors, whereas FluidGaussian largely preserves physical fidelity and interaction plausibility.
    }
    \vspace{-0.5 em}
    \label{fig:tradeoff}
\end{figure}

\vspace{-1.3 em}
\paragraph{Divergence versus Turbulence (Velocity).}
To probe robustness under stronger advection, we sweep the inflow speed
$U\in\{3,4,5,6,7,8,9,10,20,50\}$, and we plot (in Fig.~\ref{fig:tradeoff} bottom) $\Delta_{\text{gap}}=\bar{D}_{\text{Baselines}}-
\bar{D}_{\text{Ours}}$, i.e., the \emph{divergence gap}
between our method and the baseline.
To keep the initial CFL (Courant-Friedrichs-Lewy) condition fixed,
we scale the simulation timestep as
$ \Delta t_\text{sim} \;=\; \Delta t_{\text{base}}\;\frac{U_{\text{base}}}{U}$
(CFL-constant scheduling, $U_{\text{base}}=3$).
Across all tested velocities, our method maintains a positive gap,
and the advantage widens markedly in the high-turbulence regime:
when $U\in\{20,50\}$, $\Delta_{\text{gap}}$ increases sharply,
consistent with our design that penalizes compressibility in regions of strong shear. We use the Reynolds number ($Re$) to represent the level of turbulence.
The Reynolds number $Re = \frac{\rho U L}{\mu}$ measures the ratio between inertial and viscous forces in a fluid flow, where $\rho$ is the fluid density, $U$ is the flow speed, $L$ is a characteristic length (in this case, simulation domain size for simplicity), and $\mu$ is the dynamic viscosity.

\vspace{-1.3 em}
\paragraph{Physical Quality under Turbulent Flow.}
More importantly, we also evaluate fluid interactions across regimes.
As fluid velocity and $Re$ increase during testing, 
appearance-only reconstructions exhibit amplified physical errors, whereas FluidGaussian remains substantially more stable (Fig.~\ref{fig:tradeoff}, bottom).
Meanwhile, PSNR stays largely unchanged except at the highest velocities in different training settings, indicating that the physical gains do not come at a major cost in visual fidelity (Fig.~\ref{fig:tradeoff}, top).

\begin{table}[t]
\centering
\setlength{\tabcolsep}{2pt}
\captionsetup{font=small}
\caption{Visual and physics-based measures on function-critical regions of the DrivAerNet++ dataset. Adding FluidGaussian gives better reconstruction quality and lower divergence.}

\vspace{-0.7 em}

\resizebox{0.4\textwidth}{!}{
\begin{tabular}{lcccc}
\toprule
DrivAerNet++
& PSNR$\uparrow$ & SSIM$\uparrow$ & LPIPS$\downarrow$ & Divergence $\downarrow$ \\
\midrule
ActiveNeRF     & 15.6770 & 0.8843 & 0.1896 & 0.0316\\
+FluidGaussian & \textbf{16.8897} & \textbf{0.9086} & \textbf{0.1571} & \textbf{0.0231} \\
\midrule
FisherRF       & 18.2988 & 0.9206 & 0.1472 & 0.0213 \\
+FluidGaussian & \textbf{18.5436} & \textbf{0.9297} & \textbf{0.1312} & \textbf{0.0182} \\
\bottomrule
\end{tabular}
}

\vspace{-1.3 em}
\label{tab:psnr_functional_region}
\end{table}

\subsection{FluidGaussian Targets Functional Intelligence}
To evaluate the impact of FluidGaussian on the functional critical regions of each geometry, we measure the number of fluid particles interacting with each Gaussian during the SPH simulations (see Section~\ref{sec:method_function_critic}). For each geometry, the top 20 views with the highest particle interaction counts are selected from the test set. As shown in Table~\ref{tab:psnr_functional_region}, FluidGaussian consistently improves both visual and physical quality across ActiveNeRF and FisherRF on the DrivAerNet++ dataset.
FluidGaussian improves the reconstruction accuracy and reduces view divergence ($\frac{S(\mathbf{T})}{|\mathcal{G}_{\mathbf{T}}|}$), indicating that our proposed simulation-based uncertainty effectively guides view selection for physically meaningful regions.

\subsection{Ablation Study}
We decouple the influence of the fluid’s initial position by launching it from five canonical directions $\mathcal{S}_d$.
For each \(d\in\mathcal{S}_d\), we run the NBV selection driven by divergence under identical training settings and report the reconstruction metrics.
Results are shown in Table~\ref{tab:ic-ablation}.
Single-direction results vary across scenes, primarily due to the initial camera layout and occlusion patterns that determine where physically implausible regions become observable. Although the results differ under various initial conditions, all the results with FluidGaussian are consistently better than the baseline (FisherRF).

\begin{table}[!ht]

\vspace{-0.5 em}
\centering
\setlength{\tabcolsep}{2pt}
\captionsetup{font=small}
\caption{Results on Blender~\cite{mildenhall2021nerf} under different initial conditions.} %
\vspace{-0.8 em}
\resizebox{0.4\textwidth}{!}{
\setlength{\tabcolsep}{4pt}
\begin{tabular}{lccc}
\toprule
 Method (fluid direction)& PSNR$\uparrow$ & SSIM$\uparrow$ & LPIPS$\downarrow$ \\
\midrule
$\mathbf{e}_x$ & 24.3862 & 0.8857 & 0.0973 \\
$-\mathbf{e}_x$ & 24.5565 & 0.8875 & 0.0949 \\
$\mathbf{e}_z$ & 24.0015 & 0.8842 & 0.0985 \\
$-\mathbf{e}_z$ & 24.1853 & 0.8836 & 0.0986 \\
$\mathbf{e}_y$ & 24.2303 & 0.8851 & 0.0985 \\
FisherRF~\cite{jiang2023fisherrf} & 23.4230  & 0.8761  & 0.1070 \\
\makecell{FisherRF + FluidGaussian\\(all directions, Section~\ref{sec:method_nbv_selection})} 
& \textbf{24.7433} & \textbf{0.8907} & \textbf{0.0923} \\
\bottomrule
\end{tabular}

}

\label{tab:ic-ablation}
\vspace{-1 em}
\end{table}

\section{Related Works}

Recent methods improve 3D reconstruction with physical priors or simulation constraints~\cite{li2022phyir,chen2022aug, li2023pac,ni2024phyrecon,li2024neuralfluid},
while active reconstruction methods select NBVs using uncertainty or Fisher information~\cite{pan2022activenerf,jiang2023fisherrf}.
Recent uncertainty-estimation methods for radiance fields remain image-driven~\cite{lyu2024manifold, goli2024bayes}, and works on functional intelligence focus on affordances and interaction-centric representations~\cite{deng20213d,hu2015interaction,hu2016learning,hu2017learning,xiang2020sapien,han2022scene,liu2024data,delitzas2024scenefun3d,liu2025wildsmoke}.
In contrast, FluidGaussian introduces physics-aware active reconstruction by using simulation-based fluid responses to guide view selection toward function-critical regions.
Extended related work is provided in Appendix~\ref{supp:related_work}.

\section{Conclusion}
We have presented FluidGaussian, a physics-aware plug-in that injects fluid-structure diagnostics into active 3D Gaussian reconstruction. 
By re-ranking NBV candidates with simulation-derived metrics (e.g., divergence), FluidGaussian improves both visual fidelity and physical plausibility, producing simulation-ready assets and moving toward functionally intelligent reconstruction.
Future work includes: (1) We will further consider simulations of more diverse fluid properties (density, viscosity, etc.); and (2) We will consider curating and studying more non-trivial datasets of science-related geometries.

\section*{Acknowledgment}
Work at Berkeley Lab was supported by the Laboratory Directed Research and Development Program of the Lawrence Berkeley National Laboratory under U.S. Department of Energy Contract No. DE-AC02-05CH11231.
This research used resources of the National Energy Research Scientific Computing Center (NERSC), a U.S. Department of Energy Office of Science User Facility located at Lawrence Berkeley National Laboratory, operated under Contract No. DE-AC02-05CH11231 using NERSC award ASCR-ERCAPm4801.

{
    \small
    \bibliographystyle{ieeenat_fullname}
    \bibliography{main}
}

\appendix
\clearpage
\setcounter{page}{1}

\section{Experiment Details}
\label{supp:detail}
\subsection{Dataset Preprocessing}
DrivAerNet++ provides 3D geometry in STL format. To render images, we imported each object into Blender and used the BlenderNeRF~\cite{RaafatBlenderNeRF2024} extension to generate views in the same format as the Blender dataset. We set the axis-aligned bounding box to 4 and used a camera radius of 6. The first 200 rendered images were used for training or as candidate views, and another 100 were reserved for testing. The camera order was randomized. We set the background to transparent and assigned the object a gray color with hex code \texttt{\#909090}.

\subsection{Simulation Settings}
\paragraph{Fluid Initialization.}
For each experiment we select a flow direction 
$d\!\in\!\mathcal{S}_d=\{\mathbf{e}_x,-\mathbf{e}_x,\mathbf{e}_z,-\mathbf{e}_z,-\mathbf{e}_y\}$. 
Given the rigid body’s bounding box $[{\bf r}_{\min},{\bf r}_{\max}]$, an automatic fluid slab 
$[{\bf f}_{\min},{\bf f}_{\max}]$ of thickness \texttt{thick} is placed adjacent to the rigid along~$d$, 
with padding to avoid intersection and spanning the rigid in the orthogonal directions. 
The global domain $[{\bf d}_{\min},{\bf d}_{\max}]$ is enlarged accordingly and discretized with 
particle spacing equal to \texttt{particle diameter}.
All particles inside $[{\bf f}_{\min},{\bf f}_{\max}]$ are initialized as fluid particles with uniform velocity
\[
\mathbf{v}_i(0) = -\,\lambda\, d,\qquad 
\lambda=\texttt{vel},
\]
while rigid particles remain static.

\paragraph{From 3D Gaussians to Simulation Voxels.}
To bridge neural reconstruction and physics-based simulation, we convert the current 3D Gaussian representation $\mathcal{G}_t$ into a set of volumetric boundary samples that can be consumed by our SPH simulator for fluid--solid interaction handling. We compute a global axis-aligned bounding box (AABB) from all Gaussian centers and apply additional padding proportional to the object extent to avoid boundary artifacts. Instead of explicitly discretizing a dense voxel grid, we control the spatial resolution implicitly using an octree hierarchy.

\paragraph{Volumetric Sampling.}
We use the volumetric sampling operator provided by Kaolin~\cite{KaolinLibrary} to directly sample points inside the union of anisotropic Gaussian volumes. Given Gaussian centers, scales, rotations (converted to quaternions), and opacities, the sampler adaptively generates volumetric samples using an octree of fixed depth. Samples whose opacity falls below a threshold (0.3 in our implementation) are discarded. This produces a dense, geometry-aware point set that approximates the occupied volume
of the reconstructed object.
The resulting point set serves as a voxel-like occupancy representation of the solid (Section~\ref{sec:method_fluid_uncertainty_metric}).

\paragraph{Rigid–Fluid Divergence Rendering.} 
Internal particles of rigid body do not interact with fluid particles and thus always have zero divergence.
Thus, when we render the divergence of a view, we only need to consider the visible surface, without any volume rendering.

\subsection{Training Schedule}

To reflect practical scenarios where view acquisition is costly, we intentionally use biased initial views that provide only partial geometric coverage.
Following FisherRF~\cite{jiang2023fisherrf}, let $n_{\text{selected}}$ be the number of views already chosen at the current iteration. The model is trained for $n_{\text{selected}} \times 100$ iterations before selecting the next view from the candidate pool with the NBV algorithm. After the full view budget is used, the model continues training until it reaches 30k iterations.

During view selection, we follow the same overall view budget as the baseline NBV methods. Let $B$ denote the total view budget. At each step, we set the top-$K$ size to the remaining budget, i.e.,
\[
K = \text{max}\{B - n_{\text{selected}}, 2\}.
\]
Thus, the number of candidates $K$ considered by FluidGaussian decreases as more views are selected.

\section{More Results}
\label{supp:more_quantitative}
We provide additional quantitative results to further evaluate the effectiveness, generality, geometric accuracy, and efficiency of FluidGaussian.

\subsection{Random View Selection Ablation}
To isolate the contribution of our physics-aware NBV strategy, we compare random view selection and FluidGaussian under the same reconstruction baselines. As shown in Table~\ref{tab:ablation_random_vs_physics_nbv_appendix}, replacing random selection with FluidGaussian consistently improves PSNR across Blender, Mip-NeRF360, and DrivAerNet++ for both ActiveNeRF and FisherRF.

\begin{table}[h]
\vspace{-0.8em}
\centering
\captionsetup{font=small}
\caption{Comparison between random view selection and FluidGaussian under the same reconstruction baseline. PSNR is reported.}
\resizebox{0.8\linewidth}{!}{
\setlength{\tabcolsep}{3pt}
\begin{tabular}{lcccccc}

\toprule
Method & Blender & Mip & DrivAer \\
\midrule
ActiveNeRF + Random & 21.547 & 12.415 & 18.770 \\
ActiveNeRF + FluidGaussian & \textbf{21.764} & \textbf{12.479} & \textbf{18.873} \\
FisherRF + Random & 23.631 & 15.326 & 20.022 \\
FisherRF + FluidGaussian & \textbf{24.743} & \textbf{15.550} & \textbf{21.073} \\
\bottomrule
\end{tabular}
}
\label{tab:ablation_random_vs_physics_nbv_appendix}
\vspace{-1.0em}
\end{table}

\subsection{Results on ShapeNet}
We additionally evaluate FluidGaussian on 7 ShapeNet objects. As shown in Table~\ref{tab:shapenet_comp_appendix}, FluidGaussian improves PSNR and SSIM, and reduces LPIPS in both cases.

\begin{table}[h]
\centering
\captionsetup{font=small}
\caption{Comparison between the baseline and FluidGaussian on 7 ShapeNet objects.}
\resizebox{0.8\linewidth}{!}{
\setlength{\tabcolsep}{3pt}
\begin{tabular}{lcccccc}

\toprule
Method & PSNR$\uparrow$ & SSIM$\uparrow$ & LPIPS$\downarrow$ \\
\midrule
ActiveNeRF & 24.828 & \textbf{0.961} & 0.064 \\
ActiveNeRF + FluidGaussian & \textbf{25.653} & 0.960 & \textbf{0.063} \\
FisherRF & 27.604 & 0.974 & 0.041 \\
FisherRF + FluidGaussian & \textbf{27.807} & \textbf{0.976} & \textbf{0.034} \\
\bottomrule
\end{tabular}
}
\label{tab:shapenet_comp_appendix}
\vspace{-1.0em}
\end{table}

\subsection{Chamfer Distance}
We sample point clouds from the reconstructed Gaussians and from the corresponding perfect geometry (defined in Section~\ref{exp:phy_quality}), and compute the Chamfer Distance between them. As shown in Table~\ref{tab:chamfer_comp_appendix}, FluidGaussian achieves lower Chamfer Distance than both ActiveNeRF and FisherRF baselines, indicating improved geometric accuracy.

\begin{table}[h]
\centering
\captionsetup{font=small}
\caption{Chamfer Distance comparison on the Blender dataset. Lower is better.}
\resizebox{0.8\linewidth}{!}{
\setlength{\tabcolsep}{3pt}
\begin{tabular}{lc}
\toprule
Method & Chamfer Distance $\downarrow$ \\
\midrule
ActiveNeRF & 0.0103 \\
ActiveNeRF + FluidGaussian & \textbf{0.0055} \\
FisherRF & 0.0165 \\
FisherRF + FluidGaussian & \textbf{0.0046} \\
\bottomrule
\end{tabular}
}
\label{tab:chamfer_comp_appendix}
\end{table}
\vspace{-1.0em}

\subsection{Time Cost}
We also compare the running cost of FluidGaussian and the baselines. Each fluid simulation takes 1.23 minutes on average. The total cost per geometry is about 45 minutes for FluidGaussian, compared with about 17 minutes for FisherRF and 18 minutes for ActiveNeRF. Therefore, FluidGaussian is only about 2.5$\times$ overall. At approximately $64^3$ resolution, our simulator runs at 0.098 seconds per step, which is comparable to NeuralFluid~\cite{li2024neuralfluid}, which reports 0.105 seconds per step.

\section{Additional Related Works}
\label{supp:related_work}

Due to page limits, we defer our complete related works to this section.

\subsection{Physics-Informed 3D Reconstruction}
Recent advances in physics-aware reconstruction incorporate physical priors into neural representations to improve geometric accuracy and physical plausibility. Recent works~\cite{li2022phyir,chen2022aug} integrate physically grounded appearance constraints, such as reflectance modeling and illumination-consistent augmentations, to improve visual quality. Other works enforce physically plausible behavior of geometry by introducing physical simulation constraints, including continuum-material modeling, particle-based dynamics and differentiable Navier–Stokes solvers for fluid–structure~\cite{li2023pac,ni2024phyrecon,li2024neuralfluid}. 
Both of these directions enhance reconstruction quality and encourage physical consistency, but they do not consider physical interaction during the reconstruction process, and therefore they cannot shape the reconstructed geometry according to functional importance. In contrast, our method introduces physics-based metrics to prioritize the reconstruction of functionally important regions.

\subsection{Active Reconstruction}
Active view reconstruction aims to improve 3D scene recovery by adaptively selecting informative camera poses.
Building on the success of radiance-field representations~\cite{mildenhall2021nerf,kerbl20233d}, recent approaches~\cite{pan2022activenerf,jiang2023fisherrf} employ uncertainty estimation and Fisher information to guide NBV selection, effectively reducing redundancy and improving visual reconstruction quality.
However, these methods remain purely \emph{visual} in nature: their objectives optimize photometric fidelity, but they neglect the underlying physical plausibility of reconstructed geometry.
In contrast, our work introduces physics-aware active reconstruction, where view selection is informed not only by visual uncertainty but also by simulation-derived physical metrics.

\subsection{Functional Intelligence in 3D Reconstruction and Generation}

Beyond purely physics-informed priors, a complementary line of work studies 3D representations that are organized around what actions they afford. At the object level, datasets such as 3D AffordanceNet and PartNet-Mobility annotate shapes with dense affordance labels and articulation, enabling per-point functional fields instead of only class or part tags~\cite{deng20213d,xiang2020sapien}. At the scene level, SceneFun3D and functional scene reconstruction frameworks explicitly model interactive elements, motion parameters, and actionability so that the reconstructed digital twin can be operated in simulation~\cite{delitzas2024scenefun3d,han2022scene}. In graphics, ICON and follow-up work on neural functionality and functionality-focused surveys argue for descriptors built around object--object interactions and functional similarity rather than purely visual similarity~\cite{hu2015interaction,hu2016learning,hu2017learning}. Inspired by this notion of functional intelligence, our FluidGaussian approach treats simulation-derived fluid responses as functional signals and biases active reconstruction toward regions most critical for downstream fluid–object interaction.

\subsection{Uncertainty in radiance fields.}
Recent work has developed explicit uncertainty estimators for
radiance fields, including stochastic or variational formulations
such as ActiveNeRF~\cite{pan2022activenerf}, Bayes’ Rays~\cite{goli2024bayes}, and FisherRF~\cite{jiang2023fisherrf}, and more recent
manifold-based sampling methods~\cite{lyu2024manifold}. These
methods quantify epistemic ambiguity in the radiance field by
sampling or approximating distributions over appearance and
geometry parameters. While all methods have shown strong performance in image-space tasks, these uncertainty estimates remain visual-driven: they measure how well the current radiance field explains the input views or the distribution of plausible scene attributes. They do not capture whether the reconstructed geometry behaves reliably under physical interaction. 
In contrast, our FluidGaussian approach targets a different notion of reliability: simulation-propagated physical quality, which cannot be inferred from photometric evidence alone. Thus, our divergence signal complements, rather than replaces, existing uncertainty formulations.

\end{document}